\documentclass[times,twocolumn,final]{elsarticle}

\usepackage{prletters}
\usepackage{framed,multirow}
\usepackage{dsfont}

\usepackage{amssymb}
\usepackage{amsmath}
\usepackage{amsfonts}
\usepackage{latexsym}
\usepackage{breqn}
\usepackage{mathrsfs}
\usepackage{mathptmx}
\usepackage{graphicx}
\usepackage{textcomp}
\usepackage{xcolor}
\usepackage{hyperref}
\usepackage{array}
\usepackage{subcaption} 
\usepackage{url}
\usepackage{xcolor}
\definecolor{newcolor}{rgb}{.8,.349,.1}

\usepackage{algorithm}
\usepackage{algpseudocode}

\definecolor{red}{rgb}{1,0,0}
\definecolor{black}{rgb}{0,0,0}
\definecolor{purple}{rgb}{0.65,0,0.65}
\definecolor{dark_green}{rgb}{0, 0.5, 0}
\definecolor{blueish}{rgb}{0.0, 0.3, .6}
\definecolor{LightCyan}{rgb}{0.88,0.95,1}
\definecolor{tabhighlight}{HTML}{e5e5e5}

\newcommand{\rev}[1]{{\color{black}#1}}

\journal{Computer Vision and Image Understanding}

\newif\iffinal
\finaltrue
\begin{document}

\ifpreprint
  \setcounter{page}{1}
\else
  \setcounter{page}{1}
\fi

\begin{frontmatter}

\title{Spike-TBR: a Noise Resilient Neuromorphic Event Representation}
\author[1]{Gabriele \snm{Magrini}}\ead{gabriele.magrini@unifi.it}

\author[2]{Federico \snm{Becattini}\corref{cor1}}
\cortext[cor1]{Corresponding author}\ead{federico.becattini@unisi.it}

\author[1]{Luca \snm{Cultrera}}\ead{luca.cultrera@unifi.it}

\author[1]{Lorenzo \snm{Berlincioni}}\ead{lorenzo.berlincioni@unifi.it}

\author[1]{Pietro \snm{Pala}}\ead{pietro.pala@unifi.it}

\author[1]{Alberto \snm{Del Bimbo}}\ead{alberto.delbimbo@unifi.it}

\address[1]{University of Florence}
\address[2]{University of Siena}

\received{xxx}
\finalform{xxx}
\accepted{xxx}
\availableonline{xxx}
\communicated{xxx}

\begin{abstract}
Event cameras offer significant advantages over traditional frame-based sensors, including higher temporal resolution, lower latency and dynamic range. However, efficiently converting event streams into formats compatible with standard computer vision pipelines remains a challenging problem, particularly in the presence of noise. In this paper, we propose Spike-TBR, a novel event-based encoding strategy based on Temporal Binary Representation (TBR), addressing its vulnerability to noise by integrating spiking neurons. Spike-TBR combines the frame-based advantages of TBR with the noise-filtering capabilities of spiking neural networks, creating a more robust representation of event streams. We evaluate four variants of Spike-TBR, each using different spiking neurons, across multiple datasets, demonstrating superior performance in noise-affected scenarios while improving the results on clean data. Our method bridges the gap between spike-based and frame-based processing, offering a simple noise-resilient solution for event-driven vision applications.
\end{abstract}


\end{frontmatter}

%\linenumbers

% \fb{\textbf{DOMANDE E COSE CHE CI SERVONO}:
% 	\begin{itemize}
% 		\item Risultati aggiornati
% 		\item Confronto con SOTA (siamo SOTA o no?)
% 		\item Descrizione accurata del modello
% 		\item Outline esperimenti
% 		\item Confronti su metodi di denoising
% 	\end{itemize}
% }

\section{Introduction}
Traditional RGB cameras are capable of acquiring the content of a scene and output a stream of RGB frames, typically at rates of about 30 frames per second. 
When some fast-moving object is present in the scene or when the same camera is rapidly moving with respect to the scene, such a limited frame rate may yield a degradation of the sensing quality with motion blur artifacts in the acquired frames. 
Another typical limitation of traditional cameras relates to their dynamic range, i.e. the ratio between levels of the brightest and darkest parts of an image: the joint presence of very bright and dark regions can compromise the quality of the acquired scene content in some areas.
These limitations are addressed by event cameras, also referred to as neuromorphic cameras, through the use of a different image-acquiring paradigm that produces asynchronous events per pixel whenever an illumination change is detected.

Frame-based cameras capture entire images at regular time intervals, while event cameras only output information when there is a change in the visual scene. This asynchronous nature of event cameras requires adapting vision algorithms to handle these asynchronous data streams.
The simplest approach is to accumulate events over time to generate a frame-like representation. Several works use this method \cite{nguyen2019real,miao2019neuromorphic,ghosh2019spatiotemporal,cannici2020differentiable}, which is useful when working with higher-level vision algorithms that expect frames as input. The result of temporal integration can be thought of as an "event-based video", which maintains the advantages of an event camera (e.g., high dynamic range, low power consumption) while presenting the data in a format more amenable to traditional computer vision processing pipelines.

Accumulating events into frame-based representations though has its shortcomings. The accumulation time $\Delta T$ controlling the time period within which events are gathered controls the trade-off between the temporal granularity at which the data is analyzed and the number of frames that need to be processed.
If, on the one hand, small accumulation times allow computer vision pipelines to better exploit the advantages of neuromorphic sensors by capturing fine-grained movements, on the other hand, the amount of information to be processed might make real-time processing unfeasible.
Conversely, a large $\Delta T$ allows fast computation times, but can lead to information loss, as aggregating events often requires some form of quantization.

A notable approach that attempts to overcome this trade-off is Temporal Binary Representation (TBR) \cite{innocenti2021temporal}, where each pixel is created by interpreting its event history as a binary string (event/no event) and converting it to a single base-10 value.
%which compresses all the events observed within a sequence of arbitrarily small temporal intervals into a unique lossless frame-based representation.
%The approach follows the simple intuition of interpreting a stack of $N$ binary event frames as per-pixel binary numbers represented with $N$ bits, which can be easily converted to a base-10 representation without lossy quantization effects.
Despite its simplicity and effectiveness, TBR suffers from a high sensitivity to noise: the presence of noise in recent time slices can yield a flip in the most significant bit, largely changing the resulting representation.
%spurious events in the most significant digits of the binary representations can lead to relevant alterations in the values of the TBR pixels.
    % Graph-based approaches in event camera processing involve representing the spatial relationships between events using graph data structures \cite{??}. This representation facilitates efficient computations,
    % such as shortest path algorithms or topological sorting, to derive meaningful information from the event streams. 
An orthogonal strategy for processing event data involves the use of Spiking Neural Networks (SNNs) \cite{liu2021event,lee2020spike,cuadrado2023optical}, a type of artificial neural network inspired by the biological functioning of neurons in the brain. In an SNN, neurons communicate with each other via \textit{spikes} or action potentials instead of continuous signals like in traditional artificial neural networks. This theoretically makes SNNs particularly well-suited for processing event data as spikes can directly represent the events generated by an event camera, yet they do not scale well with typical synchronous hardware.
%SNNs have been extensively studied and used to process event data for various tasks, such as action recognition \cite{liu2021event}, optical flow estimation \cite{lee2020spike,cuadrado2023optical}, and object recognition\cite{?}.

Inspired by the promising properties of SNNs, we propose an enhanced version of TBR that follows a hybrid approach between the spike-based and frame-based paradigms. Our novel encoding strategy, \textit{Spike-TBR}, leverages an SNN layer to accumulate events on its membrane potential \rev{(i.e its current charging state)}, and we interpret the spikes it fires as the binary digits that make up the TBR representation.  
%In this paper we propose an encoding strategy, Spike Filtered Temporal Binary Representation, to losslessly encode the event data into a single frame given an accumulation interval while being noise-robust. A preliminary version of TBR has already been proposed in \cite{innocenti2021temporal}.
%As a first step, the event stream is processed by a Spiking Neural Network layer, that accumulates events into its membrane. The then periodically "read" the membrane to verify the presence or absence of spikes and build the TBR representation on spike trains rather than binary event sequences. 
The resulting representation is frame-based as TBR, yet it exhibits a highly increased robustness to noise due to the SNN layer, which filters noise by accumulating events into a decaying membrane and fires spikes only when the pixel-wise potential exceeds a threshold. 
We demonstrate resilience to noise even when training on clean data.

The main contributions are the following:
%\begin{itemize}
    i) we propose Spike-TBR, a noise-resistant frame-based event encoding strategy that overcomes the limitations of TBR \cite{innocenti2021temporal};
    ii) we analyze four different variants of our approach, studying different SNN neurons;
    iii) we test on four different datasets, obtaining higher or comparable results to state-of-the-art approaches.
 %A thorough analysis on the impact of Spiking Neurons on data-implicit frequencies.
%\end{itemize}

\begin{figure*}
    \centering
    \includegraphics[width=.85\linewidth]{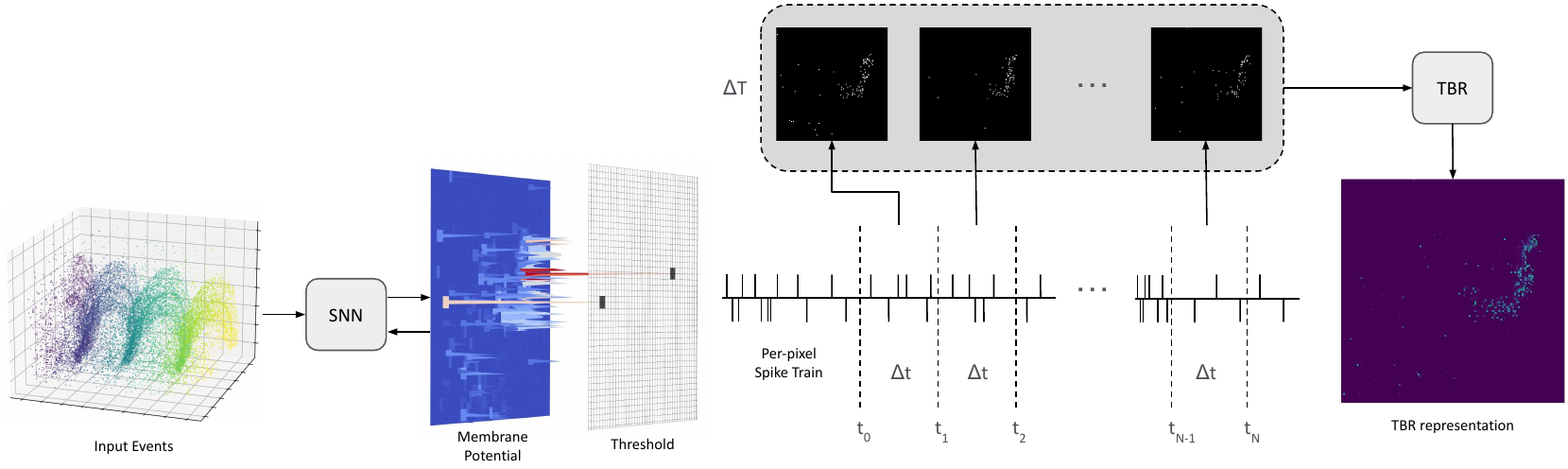}
    \caption{Spike-TBR pipeline. As the raw event stream is generated by the sensor (left) the SNN membrane accumulates potential $V(t)$. Once this potential reaches the threshold $V_{thr}$ a spike is released. This stream is constantly sampled with a $\Delta t$ period into binary frames, which every $\Delta T$ are aggregated into a TBR frame.}
    \label{fig:spikeTBR}
\end{figure*}

\section{Related Work}
\label{sec:related}

%\paragraph{Frame-based event representations}
Several works have addressed the issue of processing event camera data streams~\cite{gallego2019event}.
%Some approaches rely on specific asynchronous architectures like Spiking Neural Networks~\cite{paredes2019unsupervised, gallego2017event}, which process events one by one and incrementally store information to generate an output. However these models require specific hardware to run efficiently.
The interest to adopt traditional computer vision techniques has led to the development of a plethora of strategies to generate event representations, also in the form of event frames \cite{Gehrig-2019, cannici2020differentiable, miao2019neuromorphic, nguyen2019real, ghosh2019spatiotemporal, cannici2019asynchronous, innocenti2021temporal}.
However, most of these approaches commonly apply a temporal quantization in the form of histograms~\cite{ghosh2019spatiotemporal} or event subsampling~\cite{kaiser2019embodied}.
%\fb{to be extended adding stuff on frame-based representations}
An alternative approach to processing events is to feed them to specifically tailored architectures, such as Spiking Neural Networks (SNNs).
%\paragraph{Spiking Neural Networks}
Inspired by the fundamental mechanism of brain computation, SNNs pose themselves as the third generation of neural networks \cite{maass1997networks}, as opposed to the McCulloch-Pitts neuron \cite{mcculloch1943logical} (first generation) and feedforward and recurrent networks (second generation). The purpose of SNNs is to emulate and reach the computational efficiency of the human brain.
Although still far from complete maturity, research on SNNs has seen increasing interest in recent years, driven by the growing need for low-power-consuming neural networks. Thus, they have found applications in various empirical fields, especially when paired with event-based cameras \cite{liu2021event,lee2020spike,cuadrado2023optical}. 
%On the other hand, the sparse nature of the Spiking Neural Networks and the binary form of the propagated information makes them particularly interesting when paired with Event-Based Cameras.
To this end, various forms of integration between SNNs and neuromorphic cameras have been proposed and implemented; in \cite{lee2022fusion}, the authors present a sensor fusion framework for optical flow estimation, using both frame and event-based sensors data, leveraging the energy-efficiency of SNNs. In \cite{cordone2021learning}, a method is proposed for processing data from event cameras using spiking convolutional neural networks (SCNNs), using their sparse and asynchronous nature for efficient visual information processing. \cite{hagenaars2021self} proposes a self-supervised framework for estimating optical flow from event camera data using spiking neural networks, all in an end-to-end pipeline between event data and SNN.

%\paragraph{Hybrid Networks}
Recently, an alternative has also been investigated to take advantage of both the intrinsic temporal richness of SNNs and the computational power of Artificial Neural Networks (ANN), with growing attention for Hybrid Spiking-Artificial Networks \cite{yang2019dashnet, lee2020spike, negi2023best},  developed by implementing SNN layers as temporal encoders that feed features to an ANN decoder to estimate optical flow \cite{lee2020spike, negi2023best}.
Other approaches exploit SNNs to guide attention-based mechanisms taking into account pixel locations of spike activations \cite{yang2019dashnet}.
%propose an end-to-end self-supervised network with SNN layers acting as temporal encoding, and the ANN following containing the residual and decoder layers. \cite{negi2023best}
%[Best of Both Worlds: Hybrid SNN-ANN Architecture for Event-based Optical Flow Estimation]
%also proposes a sequentially hybrid ANN-SNN architecture for optical flow estimation. A top-down attention mechanism by incorporating an artificial neural network (ANN) to generate attention maps based on the features extracted by a feedforward SNN is presented in \cite{10.3389/fnins.2022.949142}.
Frame-based approaches inspired by SNNs have also been developed, such as the Leaky Surface~\cite{cohen2015event}, where a frame-like surface is exploited to keep track of past events: when an event occurs, the pixel at the correspondent coordinate is incremented, while the others decay through time.
This concept has also been exploited and adapted in recent works~\cite{cannici2019asynchronous}.
In this paper, the authors leverage SNN layers to ingest events asynchronously, while synchronously gathering the spikes generated for each pixel and forming a Temporal Binary Representation (TBR) \cite{innocenti2021temporal}. This offers considerable advantages compared to the original TBR formulation, which is highly noise-sensitive.
%\paragraph{Noise Filtering}
In fact, SNNs have been proven to have important signal processing properties. The Leaky Integrate and Fire (LIF) \cite{CHOWDHURY202183} declination of SNN has been shown to provide improved robustness and better generalization thanks to its high-frequency signal filtering, thus acting resistant against noisy spikes \cite{park2021noise,10.1007/978-3-031-10522-7_1}. %A linear variant of the LIF neuron has also been proposed \cite{10079519}, showing an increased level of robustness while also making the network more transparent regarding its dynamics. 
%Other works have focused on the impact of noise on Spiking Neural Networks.

\section{Event Representation}
\label{sec:method}
We define the output of an event camera with sensor size $W \times H$ as a stream of events ${x,y,t,p}$, where $x \in [1,...,H]$ and $y \in [1,...,W]$ denote the spatial coordinates of an event, $t$ the timestamp at which it occurs, and $p\in \{-1, +1\}$ its polarity, i.e., the sign of the illumination change.
We propose Spike-TBR, an aggregation strategy that groups events occurring within a predefined accumulation time $\Delta T$ into a single frame-based representation. Our solution aims to improve the Temporal Binary Representation (TBR) \cite{innocenti2021temporal}, by reformulating it using spiking layers. In the following we first provide a description of TBR, we provide the details of our proposed representation and we design a classification model exploiting Spike-TBR.

\subsection{Temporal Binary Representation}
The recent Temporal Binary Representation \cite{innocenti2021temporal} builds frame-based encodings by splitting the accumulation time $\Delta T$ into $N$ slices of equal extent $\Delta t$.
For each slice $i$, a binary frame $b^i$ is created by checking for the presence or absence of events within the timespan $[t^i; t^i + \Delta t ]$. It must be noted that, in this formulation, the polarity of events is discarded and only the spatial coordinates are considered. In addition, a single event is sufficient to activate a pixel, thus if more than one event occurs within the same temporal slice, such information is ignored.
Formally, each pixel of the binary frame $b^i$ is defined as 
%\begin{equation}
$b^i_{x,y} = \mathds{1}(x,y)$
%\end{equation}
where $\mathds{1}(x,y)$ is the indicator function returning 1 if at least an event occurred in position $(x,y)$ and 0 otherwise.
The $N$ consecutive temporal slices are then stacked together into a tensor $B \in \mathbb{R}^{H \times W \times N}$. Each spatial coordinate of the tensor can be interpreted as a channel-wise binary string with $N$ digits $B_{x,y}=[b^0_{x,y}~ b^1_{x,y}~ ...~ b^{N-1}_{x,y}]$, where the most significant digit $b^{N-1}_{x,y}$ indicates the presence of an event in the most recent temporal slice.
The final TBR representation is obtained by applying a binary-to-decimal conversion for each pixel, and normalizing the result to fit the $[0,1]$ range, by dividing it by $2^{N}-1$. Such representation condenses into a $H \times W$ tensor all the information that happened in the accumulation time $\Delta T$, losslessly up to a precision of $\Delta t$.

\subsection{Spike-TBR}\label{subsec:models}
The downside of the original TBR formulation is its high sensitivity to noise.
This stems from the fact that each pixel’s representation is derived by viewing its event history as a binary string (event/no event) and then converting this to a single base-10 value. Thus, noise in recent time slices leads to alterations in the most significant bits, resulting in drastically different base-10 representations.
\rev{In the context of SNNs, the \textit{membrane potential} of a spiking neuron is a time-dependent variable that represents the neuron’s internal state, $V(t)$ that accumulates incoming signals over time and determines whether the neuron emits a spike.}
To create a representation that is more robust to noise, we combine TBR with a Spiking Neural Network (SNN). SNNs are capable of accumulating asynchronous inputs and producing impulses (spikes), according to the membrane potential that controls the behavior of the neurons: whenever the membrane potential at a given pixel exceeds an internal threshold (which can be either fixed or learnable), a spike spatially located at such pixel is fired.
First, we process events with a SNN layer, that accumulates events into its membrane. This module then periodically "reads" the membrane potential for spike signals, accumulating spikes that occurred in a period of length $\Delta t$, rather than accumulating raw events.
The benefits of this approach are twofold: on the one hand, isolated events generated by sensor noise are filtered out, on the other hand, depending on the structure of the SNN layer, we can inject learnable parameters directly into the event encoding strategy.

As shown in Fig. \ref{fig:spikeTBR}, by periodically sampling the spike activity of the membrane potential, we can convert the per-pixel spike trains into binary pixel activations. We perform this operation for $N$ consecutive time slices and we stack the resulting binary frames. Finally, similarly to TBR, we convert the tensor to a base-10 representation and normalize the result.
The obtained Spike-TBR representation sacrifices the lossless property of TBR (up to each time slice $\Delta t$), yet makes it much more robust to noise and preserves its compactness as a single frame can summarize spatio-temporal information from the last $\Delta T = N\Delta t$ temporal interval.
\rev{In Algorithm \ref{alg} we summarize the procedure to generate Spike-TBR representations.}
We investigate multiple spiking layers, based on neurons with different temporal-dynamics. In the following, we provide an overview of the different kinds of SNN layers that we experimented.

\begin{algorithm}[t]
\caption{Spike-TBR Encoding Process}
\begin{algorithmic}[1]
\small
\Require Event Stream $E = \{(x_i, y_i, t_i, p_i)\}$, Time Window $\Delta T$, Time Slice $\Delta t$
\Ensure Encoded Frame $F$

\State Initialize number of TBR bits $N=int(\Delta T/\Delta t)$
\State Initialize Membrane Potential $V(x, y) = 0$ for all pixels
\State Initialize Spike Train $S(x, y, t) = 0$ for all the N time slices
\State Set Threshold $V_{\text{thr}}$

\For{each event $(x_i, y_i, t_i, p_i) \in E$}
    \State Update potential: 
    %\[
    $V(x_i, y_i) \gets V(x_i, y_i) + w(p_i)$
    %\]
    \If{$V(x_i, y_i) \geq V_{\text{thr}}$}
        \State Generate spike: $S(x_i, y_i, t_i) = 1$
        \State Reset membrane potential: $V(x_i, y_i) \gets V_{\text{rest}}$
    \EndIf
\EndFor

\For{each pixel $(x, y)$}
    \State Convert spike train to binary sequence: 
    \[
    B(x, y) = [S(x, y, t_1), S(x, y, t_2), ..., S(x, y, t_N)]
    \]
    \State Convert binary sequence to decimal value:
    \[
    F(x, y) = \frac{\sum_{i=0}^{N-1} 2^i \cdot S(x, y, t_i)}{2^N - 1}
    \]
\EndFor

\State \Return Encoded Frame $F$
\end{algorithmic}
\caption{\rev{Algorithm for Spike-TBR. Events are accumulated on the membrane potential of a spiking neuron, which generates a spike train for each pixel. Spikes are then accumulated into a single frame interpreting them as a binary string.}}
\label{alg}
\end{algorithm}

\textbf{\textit{Leaky Integrate and Fire (LIF)}} The first, base approach using SNN for a more robust TBR encoding consists of a single layer using Leaky Integrate and Fire (LIF) neurons \cite{maass2001pulsed}. LIF neurons can be modeled as an RC circuit in which the output discharge signal is released after a voltage threshold is reached. \rev{It can also be viewed as a system to mimic the behavior of biological neurons by integrating incoming signals over time while also leaking charge.} In particular, the  differential equation governing the evolution of the membrane potential is given by: 
\begin{equation}\label{eq:LIF}
\tau_m \frac{dV(t)}{dt} = - \left(V(t) - V_{\text{rest}}\right) + X(t)
\end{equation}
where $\tau_m$ is the membrane time constant, $V(t)$ the membrane potential at time $t$, $V_{\text{rest}}$ the membrane resting potential, and $X(t)$ is the input to neuron at time $t$.
When the membrane potential $V(t)$ exceeds a threshold $V_{thr}$, a spike is emitted and the membrane potential is scaled back to $V_{rest}$, such that:
%\fb{questa differenza (scalato o sottratto), non la facciamo mai vedere giusto? Cosa facciamo noi? Da Eq. 4 mi sembra di capire che scaliamo a $V_{reset}$ Mi pare anche di capire che la versione "sottratta" è quella di LRLIF, corretto? Nel caso toglierei il riferimento da questa sezione che confonde}
\begin{equation}
S[t] = \begin{cases} 
1 & \text{if } V[t] \geq V_{\text{th}} \\
0 & \text{otherwise}
\end{cases}
\end{equation}
where $S[t]$ is a function that denotes the presence of a spike at time $t$, and 
\begin{equation}
\text{if } V[t] \geq V_{\text{th}}, \quad V[t+1] \rightarrow V_{\text{rest}}
\end{equation}

Finally, since  we are operating at discrete time-steps, we can describe its subthreshold neural dynamics as follows:
\begin{equation}
V[t] = V[t-1] - \frac{1}{\tau_m}(V[t-1] - V_{\text{rest}}) + X[t]
\end{equation}

Note that this slightly differs from the dynamics described in Eq.~\ref{eq:LIF} as we do not decay the input. On the contrary, we raise the activation threshold to $V_{th}=1.1$ to make the neuron fire only when some information has been accumulated.
From Eq.~\ref{eq:LIF}, assuming $V_{rest}=0$ and no input $X[t]$, we can derive the decay rate of the neuron as
\begin{equation}
\label{eq:beta}
\beta=\frac{V[t]}{V[t-1]}=1-\frac{1}{\tau_m}
\end{equation}
which controls the amount of information that gets discarded at each timestep.
In this model a single layer of independent LIF neurons sits between the raw data and the TBR encoding. In this way, each pixel is connected to a single LIF neuron, which acts as a filter for noisy spikes, as also hinted in \cite{CHOWDHURY202183}. In particular, in this case the input is left without decay, while the threshold has been slightly increased so that only spikes received in a sufficiently small temporal interval are propagated along the network.

\textit{\textbf{Recurrent Synapse (RecLIF)}} \cite{yin2020effective} Similarly to the LIF approach, the layer is a 1-on-1 receptive field of LIF neurons. In this case, the output of each neuron is propagated back to itself, increasing the effect of spikes and further propagating them through time. 
This means that for an input $X[t]$, the neuron cell will get as actual input:
\begin{equation}
i[t+1] = X[t] + y[t-1]
\end{equation}

where y$[t-1]$ is the neuron cell output at time step $t-1$.

\textit{\textbf{Light Refract Synapse (LRLIF)}}  This approach presents a slight variation of the LIF model \cite{maass2001pulsed}, in which the reset mechanism changes from total voltage reset to partial inhibition (where after the output spike the internal neuron voltage is reset to 0), with only the spike threshold value subtracted after the spike is emitted from the voltage accumulated in the internal neuron, thus preserving more temporal information.
Therefore, we change the reset mechanism from: 
\begin{equation}
\text{if } V[t] \geq V_{\text{th}}, \quad V[t+1] \rightarrow V_{\text{rest}}
\end{equation} 
to a softer condition of: 
\begin{equation}
\text{if } V[t] \geq V_{\text{th}}, \quad V[t+1] \rightarrow V[t]-V_{th}    
\end{equation}

\textit{\textbf{Parametric LIF (PLIF)}}
Based on the original LIF neuron, %[ Incorporating Learnable Membrane Time Constant to Enhance Learning of Spiking Neural Networks ]
the Parametric LIF \cite{fang2021incorporating} models the Membrane Time Constant as a learnable parameter to be optimized during training rather than a hyperparameter.

\begin{figure}[t]
    \centering
    \includegraphics[width=\linewidth]{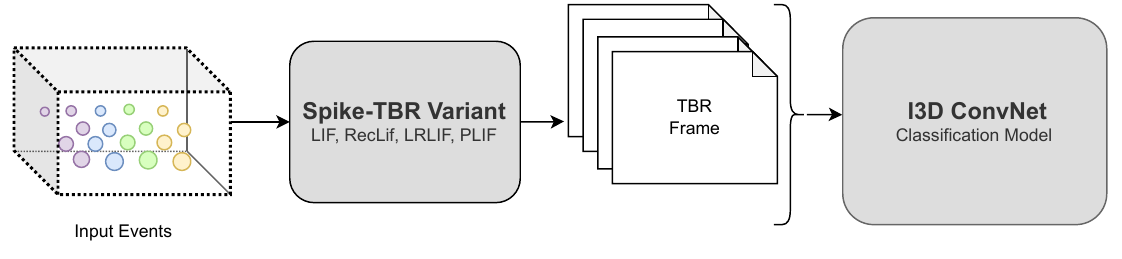}
    \caption{\rev{Pipeline experiment setup. Events are aggregated with Spike-TBR, generating a sequence of TBR frames. Then, a classifier as I3D is used.}}
    \label{fig:expi3d}
\end{figure}

% \paragraph{AdEx neuron} \todo{AdEx neuron - Da togliere?}
\paragraph{\textbf{Classification Model}}
As a classification model, we leverage an I3D ConvNet \cite{carreira2017quo} on top of our Spike-TBR frames.
Compared to \cite{carreira2017quo}, our I3D model follows a different structure. We  use a single branch Inception model with Inflated 3D convolutions. Here, the whole sequence of event frames is stacked together as a 3D tensor and processed to obtain a final 256-dimensional feature. 
%The pipeline begins with the conversion of event streams into frame-like representations using our representation method and its hybrid spike-based variations. These frames are then fed into the I3D network to perform classification.
%The key components of the pipeline are as follows:
%\begin{enumerate}
%    \item Event stream preprocessing: \todo{TBC}
%    \item Spiking Neural Network (SNN) Layer: \todo{TBC}
%    \item TBR: \todo{TBC}
%    \item I3D classification: \todo{TBC} 
%\end{enumerate}
%Unlike traditional tasks where the goal is to push the boundaries of SOTA performance, 
It must be noted that through our experiments we do not aim to achieve state-of-the-art performance on benchmarks, but rather to evaluate the effectiveness and robustness of the representation itself compared to the well-established TBR approach~\cite{innocenti2021temporal}.
In principle, any other network, specifically tailored for the task at hand, could improve the proposed results. The choice of using an I3D architecture was driven by its success in several event-based tasks in prior works \cite{innocenti2021temporal, berlincioni2023neuromorphic, berlincioni2024neuromorphic, becattini2024crossmodal}.
Our primary focus, however, is evaluating the quality and robustness of the Spike-TBR, especially under noisy conditions. \rev{The overall pipeline is shown in Fig.\ref{fig:expi3d}.}
%While I3D is commonly used for high-performance video recognition tasks, we employ it here as a tool to measure how well the generated representations preserve meaningful temporal and spatial information.
%Our analysis seeks to understand how effectively the representations generated by our hybrid approach capture event dynamics, rather than achieving the highest possible classification accuracy.

%\section{Datasets}

\begin{table*}[t]
    %\caption{Global caption}
    \begin{minipage}{.26\linewidth}
      \caption{DVSGesture-128 Results.}
      \label{tab:sota_DVS_Gesture}
      \centering
        \begin{tabular}{l|c}
			Model & Acc\\ \hline \hline
%ALERT \cite{martin2024alert} & 96.2 \\
OTT \cite{xiao2022online} & 96.88 \\
DECOLLE \cite{kaiser2020synaptic} & 97.50 \\
Event-SSM \cite{schone2024scalable} & 97.70\\
EGRU  \cite{subramoney2022efficient} & 97.80 \\
ST-SNN \cite{she2022sequence} & 98.00 \\
TRIP \cite{arjmand2024trip} & 98.60 \\ 
ACE-BET \cite{liu2022fast} & 98.88 \\ \hline		
TBR \cite{innocenti2021temporal} & \textbf{99.62} \\
\hline
Spike-TBR$_{LIF}$ & \textbf{99.62} \\
Spike-TBR$_{RecLIF}$ & \textbf{99.62} \\
Spike-TBR$_{LRLIF}$ & 99.24 \\
Spike-TBR$_{PLIF}$ & \textbf{99.62} \\
			
		\end{tabular}
    \end{minipage}%
    \begin{minipage}{.26\linewidth}
      \centering
        \caption{DVSLip Results.}
        \label{tab:sota_DVS_Lip}
        \begin{tabular}{l|c}
			Model & Acc\\ \hline \hline
S-MSTP \cite{bulzomi2023end} & 60.20 \\
%SNN-STAB \cite{liu2024intelligent}  &  63.6 \\
Rn-net \cite{yoo2024rn}  &  67.50 \\ 
Action-net \cite{wang2021action}  &  68.80 \\
G2N2 \cite{mesquida2023g2n2}  &  69.40 \\
MSTP \cite{bulzomi2023end} & 72.10 \\
MTGA \cite{zhang2024mtga}  &  75.08 \\
SpikGRU2+ \cite{dampfhoffer2024neuromorphic} & 75.30 \\ \hline
TBR \cite{innocenti2021temporal} & 70.00 \\
\hline
Spike-TBR$_{LIF}$ & \textbf{75.91} \\
Spike-TBR$_{RecLIF}$ & 74.45 \\
Spike-TBR$_{LRLIF}$ & 71.53 \\
Spike-TBR$_{PLIF}$ & 74.45\\
		\end{tabular}
    \end{minipage} 
    \begin{minipage}{.26\linewidth}
      \centering
        \caption{NCaltech-101 Results.}
        \label{tab:sota_NCaltech101}
        \begin{tabular}{l|c}
Model & Acc\\ \hline \hline
TEBN \cite{duan2022temporal} & 63.13 \\
BackEISNN \cite{zhao2022backeisnn} & 65.53 \\
ST-SNN \cite{she2022sequence} & 71.20 \\
Spikformer \cite{zhou2022spikformer} & 72.83 \\
SSNN \cite{ding2024shrinking} & 77.97 \\
NDA-SNN \cite{li2022neuromorphic} & 78.20\\
Eventmix \cite{shen2023eventmix} &\textbf{79.47} \\ \hline
                TBR \cite{innocenti2021temporal} & 72.50 \\ 
			\hline
			Spike-TBR$_{LIF}$ & 68.30 \\
			Spike-TBR$_{RecLIF}$ & 69.17 \\
			Spike-TBR$_{LRLIF}$ & 68.08 \\
			Spike-TBR$_{PLIF}$ & 69.28\\
		\end{tabular}  
  ~\\
  ~\\
    \end{minipage} 
     \begin{minipage}{.2\linewidth}
      \centering
      %MICC-G per non far andare a capo
        \caption{MICCGesture results.}
        \label{tab:sota_MICC_Gesture}
        \begin{tabular}{l|c}
			Model & Acc\\ \hline \hline
                Polarity \cite{innocenti2021temporal} & 68.40 \\
                SAE \cite{innocenti2021temporal} & 70.13 \\ \hline
			TBR \cite{innocenti2021temporal} & \textbf{73.16} \\
			\hline
			Spike-TBR$_{LIF}$ & \textbf{73.16} \\
			Spike-TBR$_{RecLIF}$ & 66.66 \\
			Spike-TBR$_{LRLIF}$ & 64.06 \\
			Spike-TBR$_{PLIF}$ & 66.23 \\
		\end{tabular}  
  ~\\
  ~\\
  ~\\
  ~\\
  ~\\
  ~\\
    \end{minipage} 
\end{table*}

\section{Experiments}
%In this section, we present the experimental results of our proposed approach across various datasets, comparing it with state-of-the-art methods.

We test our approach on 4 event classification datasets.
%This collection of datasets is composed of data collected using different cameras at different resolutions for different tasks. At the same time all them can be reduced to a classification task.
%\paragraph{DVSGesture}
The \textit{DVSGesture-128} \cite{amir2017low} 
%\todo{The dataset contains a total of 1342 hand gestures
%with a variable duration spanning from approximately 2 to 18
%seconds (6 seconds on average). Gestures are divided in 10
%classes plus an additional random class for unknown gestures.
%Each of these actions are performed by 29 subjects under
%different illumination conditions (natural, fluorescent and led
%lights). The data is acquired using a DVS128 camera, i.e. an
%event camera with a sensor size of 128 × 128 pixels.}
is composed of 1342 hand gestures, with an average duration of 6 seconds. Samples are split into 11 classes, 10 gestures plus a distractor. The recordings comprise 29 subjects under different illumination conditions. and were made using a DVS128 camera (128×128px resolution).
%\todo{SOTA for DVSGesture \cite{she2022sequence} 97.8}
%\paragraph{MICCGesture}
The \textit{MICCGesture} dataset \cite{innocenti2021temporal} was recorded by 7 different actors of different age, height and gender for a total of 231 videos. It was designed as an extension of DVSGesture-128, and comprises the same gestures at different speeds. It was recorded using a higher resolution of 640×480.
%\todo{MICC Gesture SOTA \cite{innocenti2021temporal} 73.16}
%\paragraph{DVSLip}
The \textit{DVSLip} dataset \cite{tan2022multi} is a collection of event-based lip-reading videos recorded with a DAVIS346 camera (346×260px resolution) from 40 subjects reading sequences of words.
%The recordings have been performed using a DAVIS346 camera with a spatial resolution of 346×260.
%\todo{{SOTA dvslip} \cite{dampfhoffer2024neuromorphic} 75.3}
Finally, the \textit{NCaltech-101} dataset \cite{ncaltech101} is a neuromorphic conversion of Caltech101~\cite{caltech101} obtained by recording the original data with an event camera. It is a classification task with 100 classes.
%\todo{SOTA \cite{ding2024shrinking} 77.97 VG 67.51 ResNet }
The results are shown in two sets of experiments: comparison with state-of-the-art models and robustness of the model under noisy conditions.
For all the experiments we use 8 bits for TBR and, depending on the dataset, we use slightly different settings for the accumulation time. We use a $\Delta t=2.5ms$ for DVSGesture-128, NCalTech-101 and MICCGesture, while $\Delta t=6.25ms$ for DVSLip, as events streams are sparser.
\rev{The choice of $\Delta t$ is based on the characteristics of each dataset. For datasets with denser event streams (DVSGesture-128, MICCGesture, NCaltech-101), we use a smaller accumulation time to retain finer temporal details, which might otherwise be lost. Conversely, for DVSLip, which has sparser event streams, we increase the accumulation time to prevent excessive sparsity in the encoded frames. Sparsity is a direct consequence of how subjects move in the recorded scene, as fast movements yield a much higher event rate (thus, a dense stream), whereas slower movements, as found in DVSLip, result in sparser streams as less events are generated.}
Similarly, we consider non-overlapping chunks of 500ms as separate samples at training time on DVSGesture-128, NCalTech-101 and MICCGesture, yet for DVSLip we use entire sequences instead. At test time we use chunks of the same lengths and perform a majority voting for the final classification.
\rev{All the Spike Neural Network experiments have been implemented using the \texttt{spikingjelly} \cite{jelly} and the \texttt{tonic} \cite{tonic} libraries.}
%------------- PARAMETRI
% per ogni bit
%$\Delta t=2.5ms$, $\Delta T=20ms$ DVSGesture, NCatlTech, MICCGesture
%6.25ms DVSLips (+150\% rispetto a 2.5m)
% random crop temporale di 500ms a sequenza per addestrare su DVSGesture e NCaltech. A test time crop di 500ms non sovrapposti e votazione a maggioranza (come undri). Sequenze intere su DVSLip.
%\todo{VEDI COMMENTI NEL SORGENTE}

\subsection{State-of-the-art Comparison}
%We conduct several experiments on benchmark event-based datasets, comparing our model with the top-performing approaches in the field.
In Tabs. \ref{tab:sota_DVS_Gesture}, \ref{tab:sota_DVS_Lip}, \ref{tab:sota_NCaltech101} and \ref{tab:sota_MICC_Gesture} we report the accuracies of our approach compared to state of the art methods. All results are taken from publicly available benchmarks, whereas we run all TBR \cite{innocenti2021temporal} experiments with our pipeline to establish a fair comparison.
For all the LIF variants, except PLIF which has learnable parameters, we tune the decay rate $beta$ on a held-out validation set. We use $\beta=0.5, 0.9, 0.7, 0.7$, respectively for DVSGesture-128, DVSLip, NCaltech-101 and MICCGesture.

%--------------------------------------DVSGESTURE-128----------------------------------
On DVSGesture-128 (Tab. \ref{tab:sota_DVS_Gesture}), we achieve an accuracy of 99.62\% with most models, on par with the vanilla TBR representation \cite{innocenti2021temporal}.
% The LIF neuron model equals the performance of TBR \cite{innocenti2021temporal}, highlighting that even without recurrent dynamics, LIF can effectively handle event data.
It must be said that the DVSGesture-128 dataset is simple and saturated, however, the results confirm that Spike-TBR is capable of matching or improving upon other state-of-the-art approaches.
%Models such as DECOLLE \cite{kaiser2020synaptic} and ST-SNN \cite{she2022sequence} also perform competitively, but our models consistently outperform them. %This table clearly establishes that the incorporation of spiking neurons, especially the recurrent variant, significantly improves performance on gesture recognition tasks.
%--------------------------------------DVSLip----------------------------------
Tab. \ref{tab:sota_DVS_Lip} reports the results on DVSLip~\cite{tan2022multi}. 
Interestingly, all variants of Spike-TBR yield improved results compared to TBR. In particular, the LIF neuron is able to obtain state-of-the-art results, followed closely by PLIF and surpassing previous best-performing methods such as MTGA \cite{zhang2024mtga} and SpikGRU2+ \cite{dampfhoffer2024neuromorphic}.
%The PLIF neuron model achieves a close-to-best performance of 74.45\%, just slightly behind the top-performing SpikGRU2+ \cite{dampfhoffer2024neuromorphic} (75.3\%) and MTGA \cite{zhang2024mtga} (75.08\%).
%The LIF and RecLIF neuron models have slightly lower performance on this benchmark compared to more advanced architectures like SpikGRU2+, which likely benefits from more complex temporal dynamics. %Despite this, our models remain competitive, especially given the challenging nature of the task. 
The results highlight that our proposed models maintain strong accuracy even in tasks requiring fine-grained motion understanding, such as lip movements.
%\fb{Va capito che risultati mettere e va discusso il risultato Spike-TBR vs TBR}
%--------------------------------------NCALTECH101----------------------------------
Tab. \ref{tab:sota_NCaltech101} shows the classification accuracy on NCaltech-101 \cite{ncaltech101} dataset.
On this benchmark, Eventmix \cite{shen2023eventmix} and NDA-SNN \cite{li2022neuromorphic} models, which leverage intensive data augmentation strategies, show the highest accuracies, ranking higher than all Spike-TBR variants.
However, we obtain slightly lower results compared to vanilla TBR, showing that the spiking layers can be introduced in the encoding strategy without severe side effects.
It must be noted that NCaltech-101 is the only dataset where we report lower results than TBR. We attribute this to the fact that the dataset is collected by recording a screen and events appear in bursts corresponding to the saccadic movements performed to make the signal visible to the camera. This unnatural stream of events makes it hard for spiking neural networks to accumulate information in the membrane correctly.
%Our PLIF and RecLIF approaches achieve a results of 69.28\% and 69.17\%, outperforming models like BackEISNN \cite{zhao2022backeisnn} and TEBN \cite{duan2022temporal}.
%However, the performance gap between our models and top performers like Eventmix suggests that more advanced data augmentation strategies could further improve our results.
%\todo{questa affermazione è dovuta al fatto che i due paper che EventMix e NDA-SNN sono paper che presentano tecniche di data augumentation}. \lb{Spieghiamo, di nuovo, che è un dataset strano fatto registrando un monitor ergo il nostro drop di performance}
%\fb{Va capito che risultati mettere e va discusso il risultato Spike-TBR vs TBR}
%--------------------------------------MICC-GESTURE----------------------------------
Finally, in Tab. \ref{tab:sota_MICC_Gesture} we report a comparison with state-of-the-art methods on the MICCGesture dataset. Spike-TBR$_{LIF}$ matches the accuracy of TBR, confirming the quality of the event encoding strategy.
% The RecLIF and PLIF neuron models perform slightly lower with accuracy of 66.66\% and 66.23\% respectively. However, all proposed models demonstrate strong overall performance, particularly in terms of their ability to handle complex gesture sequences.
The experiments on the four datasets show that Spike-TBR consistently provides comparable or superior results to TBR, setting a strong foundation for our subsequent analysis in noise-affected environments.
\rev{In general, these improvements are primarily due to the combination of frame-based and event-based encodings and to the noise-resistant properties introduced by spiking neurons. Additionally, the usage of learnable parameters (PLIF variant) can further optimize event aggregation.}
%\todo{Conclusioni esperimetno - Da rivedere:} Overall, these experiments demonstrate that our hybrid models leveraging Spiking Neural Networks consistently achieve competitive or superior performance across different event-based tasks, both in terms of gesture recognition and object classification.
%Our spiking neuron-based approach demonstrate competitive performance particularly when compared to TBR \cite{innocenti2021temporal}.%, which represents the same overall approach but without the inclusion of Spiking Neural Networks. 
%\todo{Indeed, our method either match or outperforms TBR \cite{innocenti2021temporal} non qui, mettere quando parliamo di noise}. The use of spiking neurons adds a significant benefit in tasks that require capturing fine temporal details, as seen in gesture recognition and lip reading. While TBR \cite{innocenti2021temporal} performs slightly better in static object classification, our models offer comparable performance, setting a strong foundation for subsequent analysis in noise-affected environments.

\newcommand{\figreswidth}{.5\linewidth}
\begin{figure}[t]
	\begin{center}
        \setlength{\tabcolsep}{2pt} 
		\begin{tabular}{cccc}
			\includegraphics[width=0.22\linewidth]{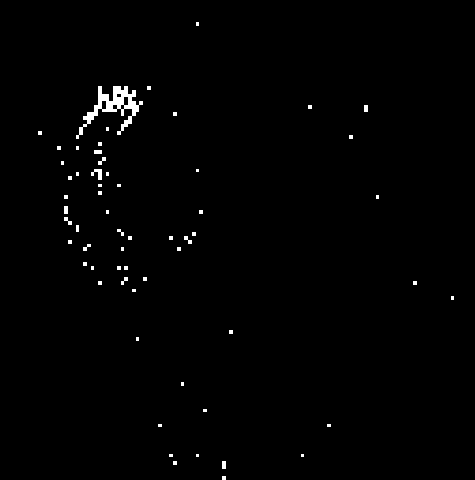} &\includegraphics[width=0.22\linewidth]{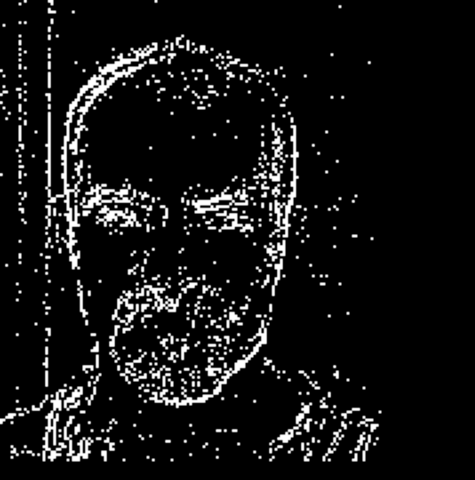}& \includegraphics[width=0.22\linewidth]{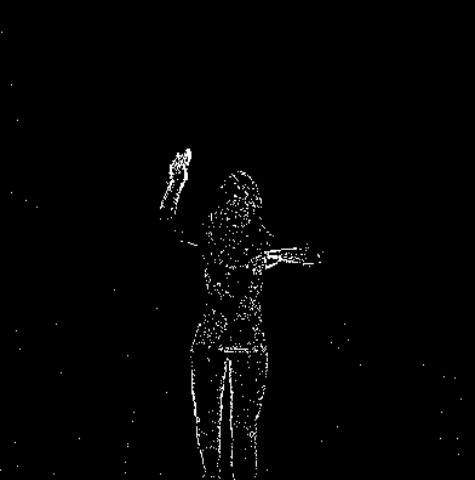}&
   \includegraphics[width=0.22\linewidth]{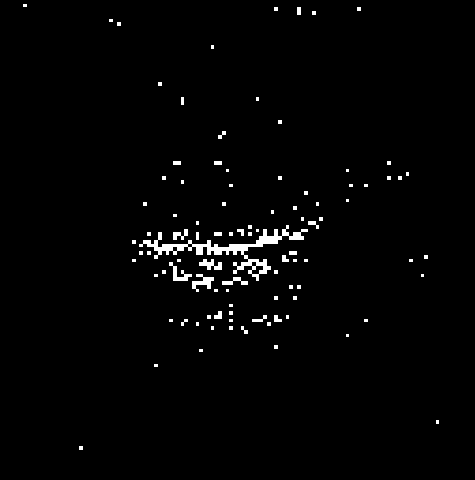}\\
			\includegraphics[width=0.22\linewidth]{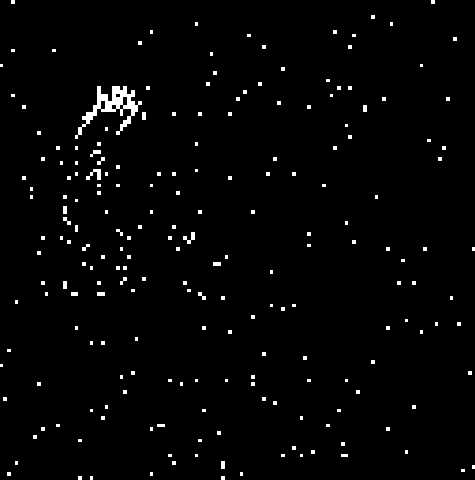} &  \includegraphics[width=0.22\linewidth]{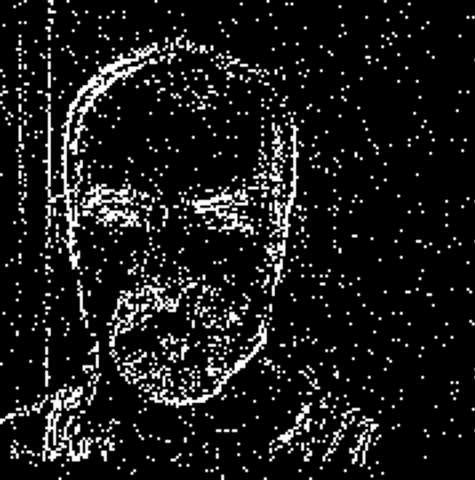} & \includegraphics[width=0.22\linewidth]{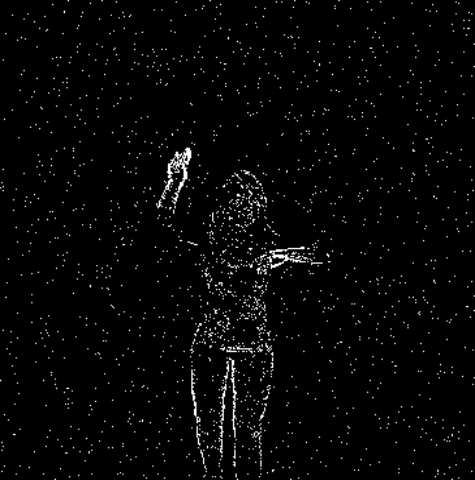} & \includegraphics[width=0.22\linewidth]{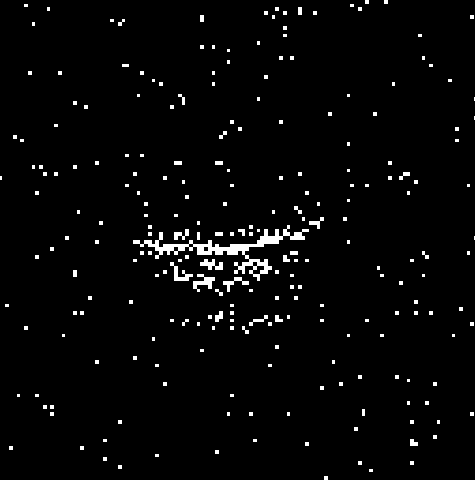} \\
			\includegraphics[width=0.22\linewidth]{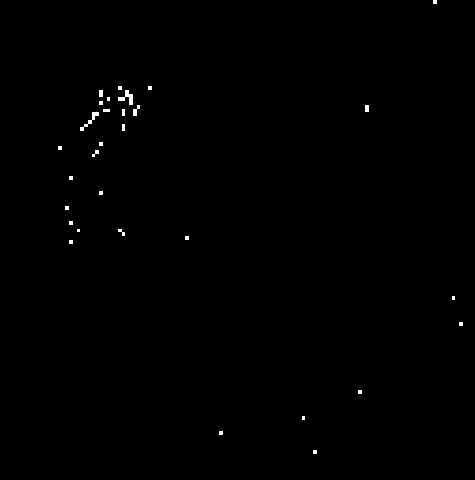} &  \includegraphics[width=0.22\linewidth]{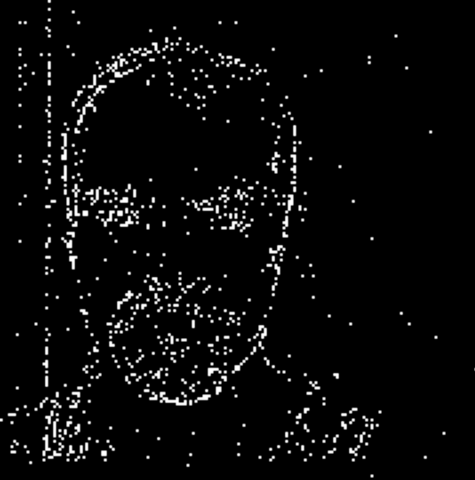} & \includegraphics[width=0.22\linewidth]{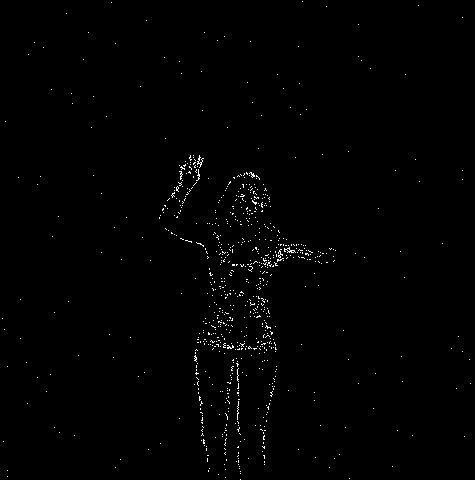} & \includegraphics[width=0.22\linewidth]{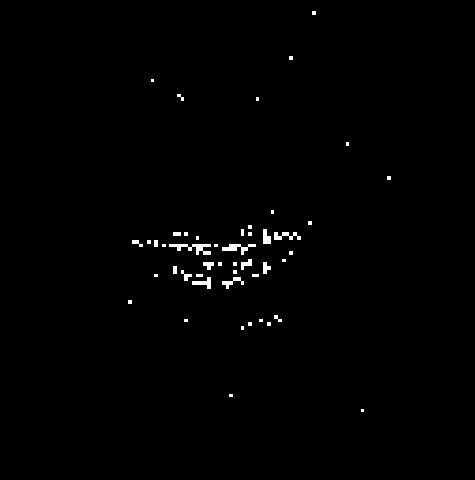} \\
		\end{tabular}
	\end{center}
	\caption{Impact of noise on binary event frames and filtering properties of Spiking Neurons. \textbf{Top}: consecutive frames extracted directly from the four datasets used in this paper. \textbf{Center}: the impact of 1\% noise on the same frames. \textbf{Bottom}: the output of the LIF receptive field using the noisy images as input.}% The LIF layer filtered through most of the additional event-noise, partially restoring the original figure event distribution.}
	\label{fig:qualitative}
\end{figure}

\begin{figure*}[t]
	\begin{center}
		\begin{tabular}{cc}
			\subcaptionbox{ DVSGesture-128 \label{subfig:dvsgesture}}{
			\centering
			\includegraphics[width=.45\linewidth]{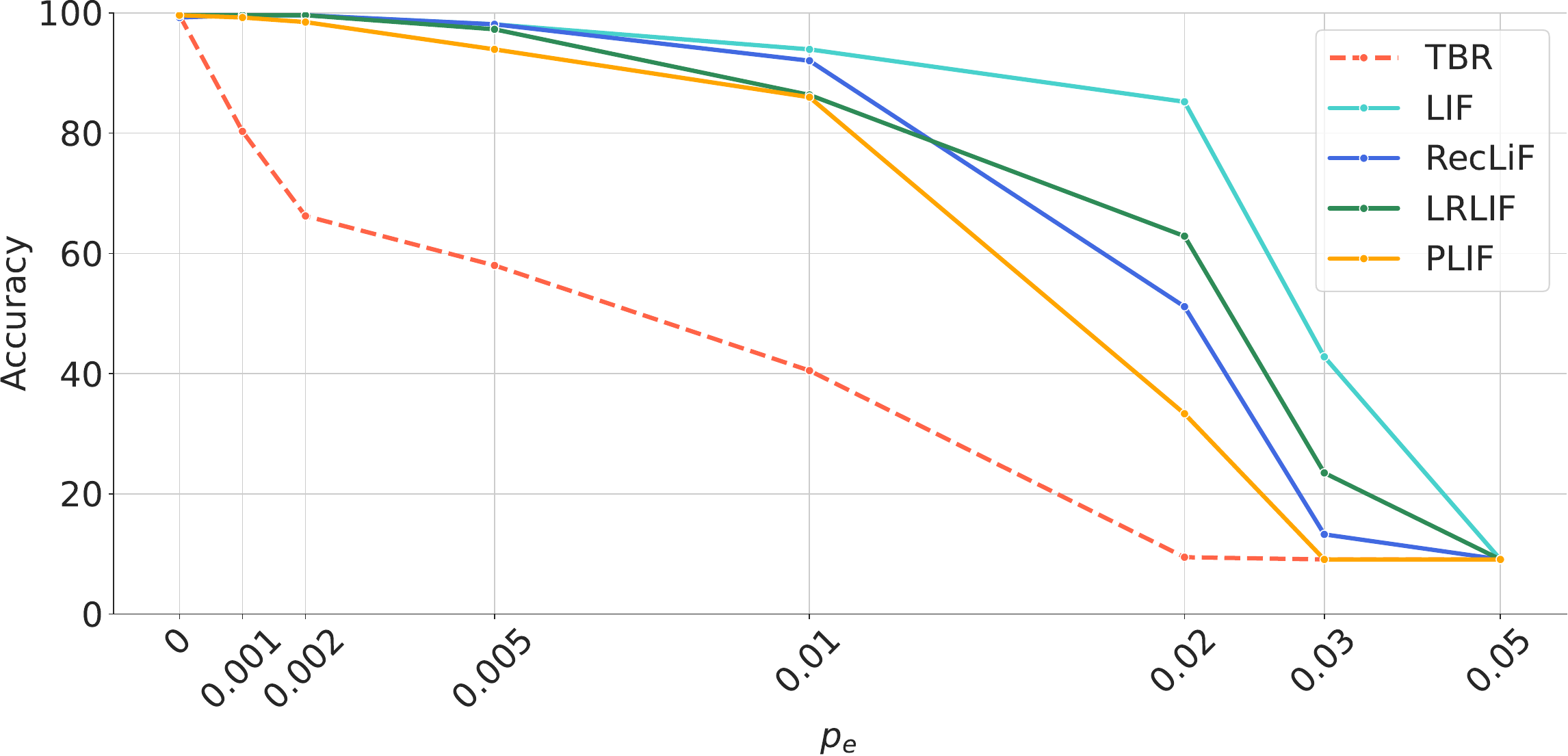}} &
			\subcaptionbox{ MICCGesture\label{subfig:miccgesture}}{
			\centering
			\includegraphics[width=.45\linewidth]{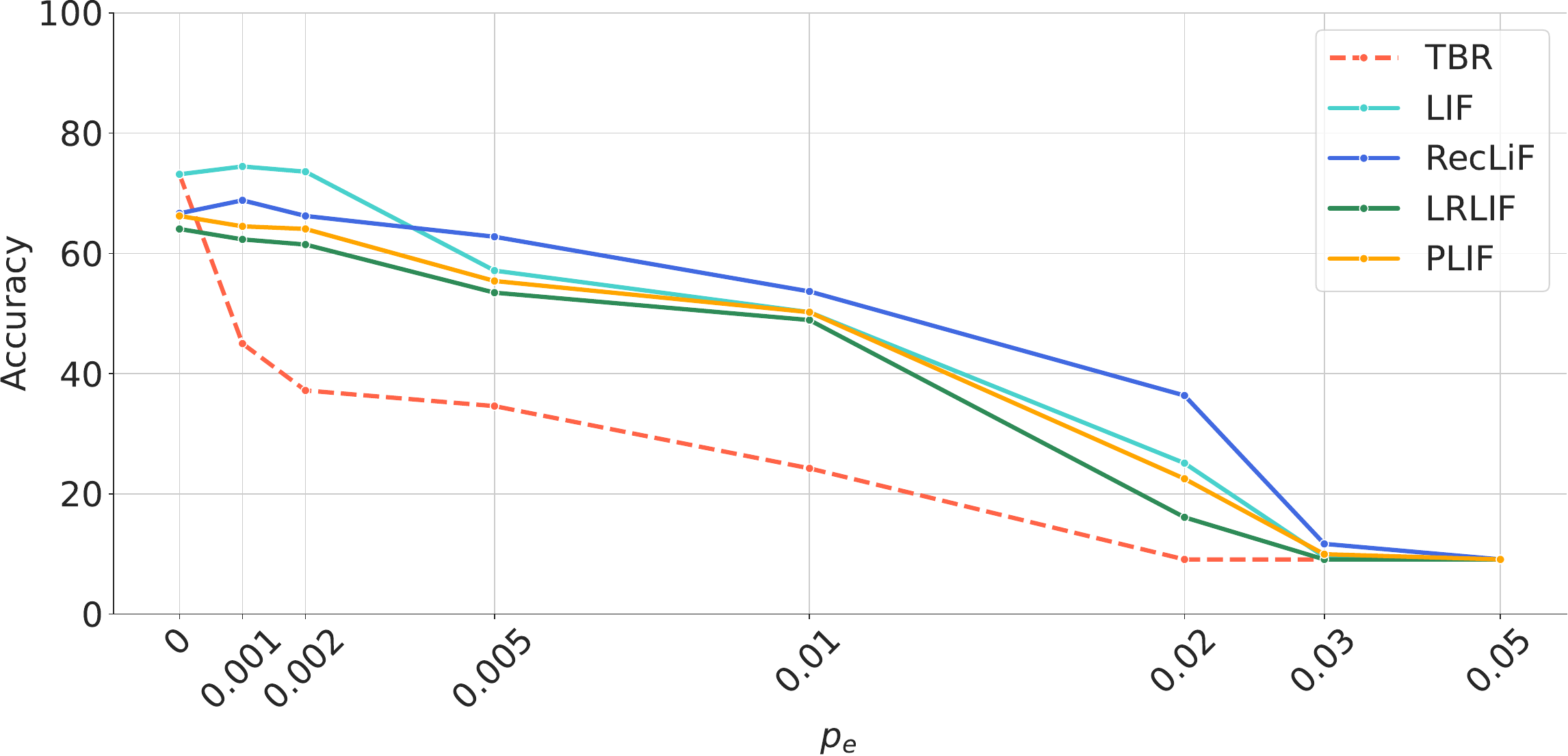}} \\
			\subcaptionbox{ DVSLip\label{subfig:dvslip}}{
			\centering
			\includegraphics[width=.45\linewidth]{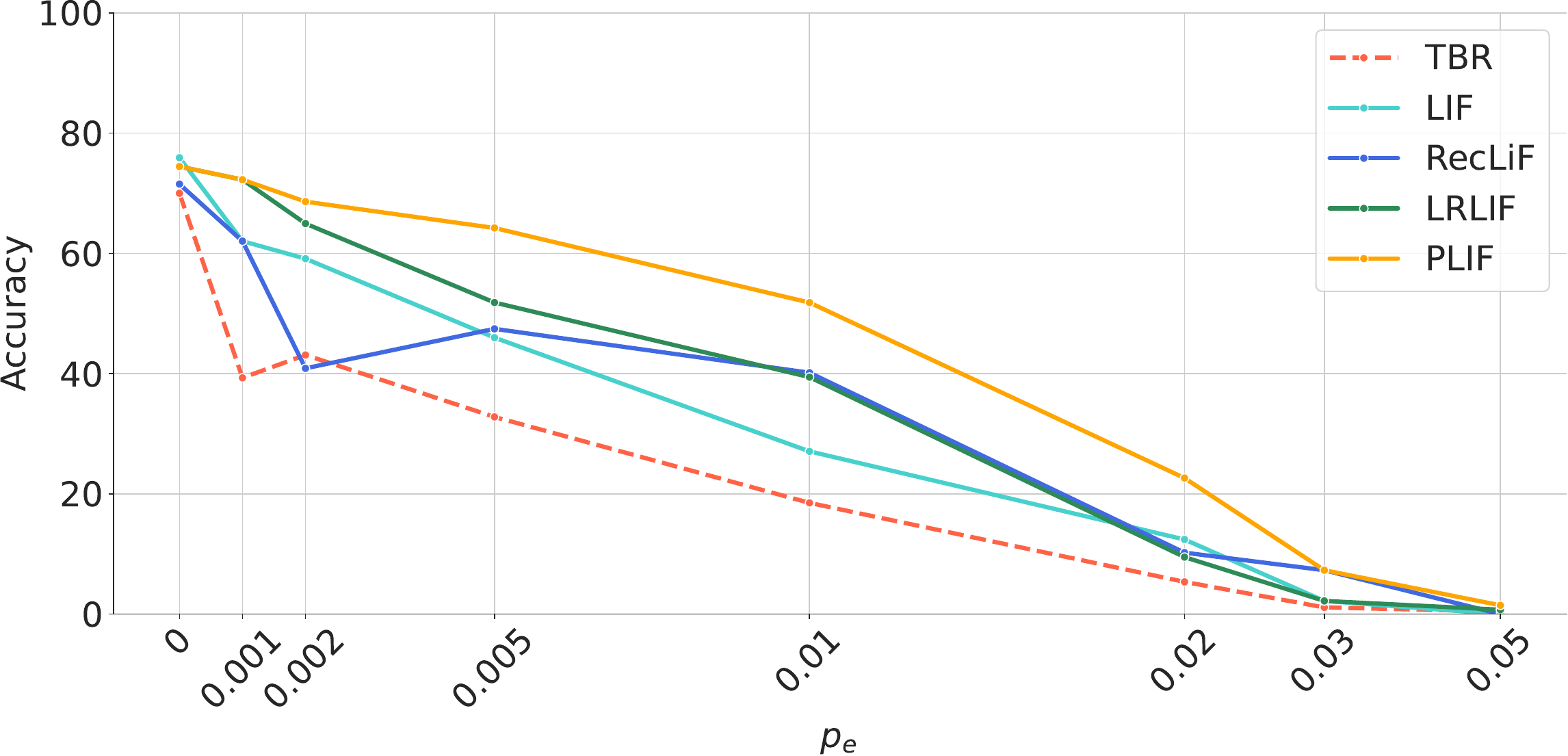}} &
			\subcaptionbox{  NCaltech-101\label{subfig:ncaltech}}{
			\centering
			\includegraphics[width=.45\linewidth]{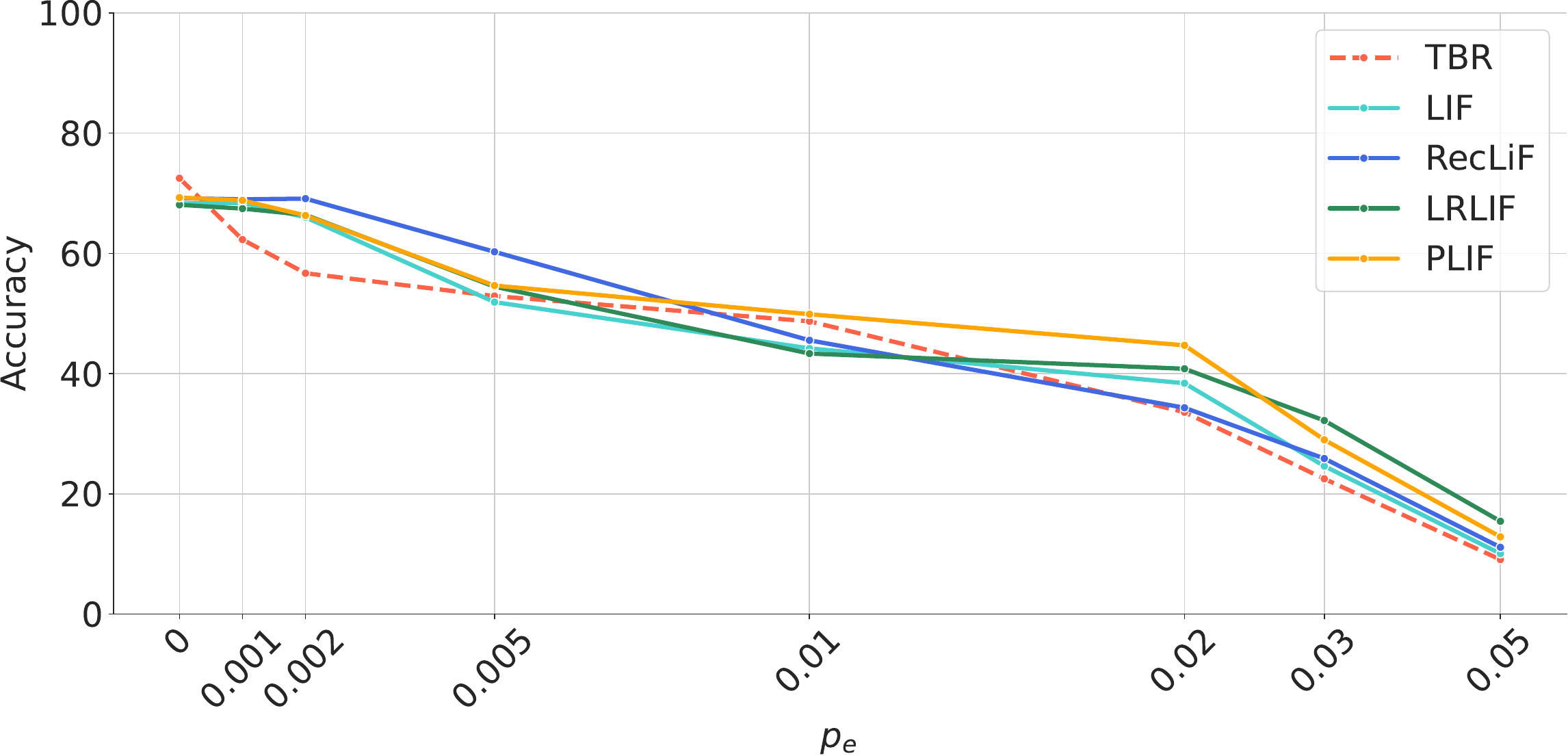}} \\
		\end{tabular}
	\end{center}
	\caption{Noise-affected performances on the tested datasets based on the different spiking neuronal architecture. Plots (a), (b), (c), and (d) correspond to DVSGesture-128, MICCGesture, DVSLip, and NCaltech-101, respectively. In all the instances, the augmented resilience with respect to the accuracy is clearly noticeable in the architectures involving a Spiking Neuron as noise filtering, compared to the catastrophic impact of the noise on the base I3D network.}
	\label{fig:detres}
\end{figure*}

\subsection{Sensitivity to Noise}
To assess the robustness to noise of Spike-TBR compared to TBR \cite{innocenti2021temporal}, we inject noise into the event streams.
%In this section, we assess the robustness of our proposed model by introducing synthetic noise into the event-based data streams.
%The primary goal of this experiment is to evaluate the ability of our model to maintain performance in the presence of varying level of noise, comparing them to the baseline TBR \cite{innocenti2021temporal} approach. 
%Given an event stream of temporal extent $\Delta t$, we randomly generate synthetic events with probability $p_e$, meaning that the resulting binary frame, accumulated over such time interval, will have a pixel activation regardless of the presence of an event with probability $p_e$.
\rev{To simulate a given amount of noise p (say $p=0.01$ for 1\%), we add an event every $\Delta t$ ms with probability $p$ for each pixel. Therefore, each TBR slice has, on average, $p \times H \times W$ noise events, where H and W are the height and width of the frame. Since a TBR frame is composed of N slices, noise injection can alter up to $p \times H \times W \times N$ pixels in the final representation.}
This setup allows us to analyze the resilience of our spiking-based models, which are designed to filter out noise and fire only if the membrane reaches sufficient potential, as shown in Fig. \ref{fig:qualitative}, where binary frames generated by Spike-TBR$_{LIF}$ are shown.
%the spiking layers are capable of filtering out noise, preserving the original signal.

Fig. \ref{fig:detres} shows the impact of increasing noise levels on classification accuracy. %The x-axis in each plot represents the incremental noise levels, while the y-axis shows the corresponding accuracy.
Across all datasets, Spike-TBR significantly enhances the robustness to noise compared to vanilla TBR.
On DVSGesture-128 (Fig. \ref{subfig:dvsgesture}) the accuracy of TBR declines rapidly as noise increases, whereas all variations of Spike-TBR maintain a higher accuracy even for large quantities of noise.
Similar trends are observed on MICCGesture (Fig. \ref{subfig:miccgesture}),
although overall accuracies are lower due to the higher complexity of the dataset and all models do not tolerate well large amounts of noise.
%where TBR shows a pronounced decline in accuracy with increasing noise. In contrast, our LIF, RecLIF, LRLIF and PLIF models show more stability, indicating that spiking neurons help filter out noise while retaining the essential gesture information.
For DVSLip (Fig. \ref{subfig:dvslip}), which involves more subtle lip movements, the gap between TBR and Spike-TBR is still evident, although there is a more appreciable difference between different spike neurons. In particular, the PLIF neuron, which has learnable parameters, outperforms all other variants.
%As noise increases, TBR struggles to maintain accuracy, while our PLIF model proves to be the most resilient, consistently outperforming TBR across all noise levels.
%This highlights the strength of our spike-based approach in capturing finer motion details even under noisy conditions.
Finally, the NCaltech-101 dataset (Fig. \ref{subfig:ncaltech}) poses a more challenging task due the presence of saccadic movements that make noise filtering harder. However, RecLIF and PLIF show greater robustness, maintaining a consistently higher accuracy across noise levels compared to TBR. 
In conclusion Fig. \ref{fig:detres} shows that, across all datasets, the spike-based approaches (LIF, RecLIF, LRLIF and PLIF) outperform TBR in noisy conditions.

\begin{table}
    \centering
    \caption{\rev{Effect of training TBR with noise injection at training time. The model proves to be more robust to the specific amount of noise, but it quickly degrades when more noise is added. Spike-TBR instead demonstrates to be more robust.}}
    \resizebox{.75\columnwidth}{!}{
    \begin{tabular}{l|ccccc}
         %\multicolumn{1}{c}{}& \multicolumn{5}{c}{DVSGesture-128} \\ \hline
         Noise & 0.0 & 0.001 & 0.01 & 0.02 & 0.05 \\ \hline
         TBR &  \textbf{99.62} & 80.30 & 40.50 & 9.46 & \textbf{9.09} \\
         TBR + noise aug. (0.001) & \textbf{99.62} & \textbf{99.62} & 11.74 & 9.09 & \textbf{9.09} \\
         %TBR + noise aug. (0.005) & 96.59 & 96.59 & 11.74 & 9.09 & 9.09 \\
         TBR + noise aug. (0.01) & 98.86 & 98.48 & \textbf{99.24} & 9.46 & \textbf{9.09} \\
         %TBR + noise aug. (0.05) & 67.42 & 67.80 & 78.03 & \textbf{84.84} & \textbf{98.86} \\
         Spike-TBR$_{LIF}$ & \textbf{99.62} & \textbf{99.62} & 93.94 & \textbf{42.80} & \textbf{9.09} \\
    \end{tabular}
     }
    \label{tab:noisetrain}
\end{table}

\rev{We perform an additional control study, by training TBR non-spiking models with noise injected at training time to establish the effect of a simple data augmentation strategy. The model indeed improves its robustness when the amount of noise used at training time is equivalent to the one added at test time. However, with higher noise, the model quickly degrades. We show in Tab. \ref{tab:noisetrain} the results of training TBR with noise 0.001 and 0.01.
This behavior occurs since the model overfits such noise levels, failing to generalize to higher ones. Moreover, the model degrades in performance also in the non-noisy scenario depending on the quantity of noise seen in training. Spike-TBR instead offers a robust alternative without prior knowledge of the noise level that will be encountered at test time.
}

\rev{It must be noted that in all our experiments we used synthetic random noise. However, other types of noise (sensor noise, thermal noise, pixel defects,  etc.) might be distributed differently. Yet, modeling these sources of noise is not straightforward.
To test on realistic noise, we first recorded a static background to gather real noise generated by an event camera\footnote{Prophesee Evaluation Kit 4 (EVK4), with IMX646 neuromorphic sensor}. Then, we added the noise to the original event videos, resizing them to match the dataset's frame size. We show in Tab.~\ref{tab:realnoise} results on the DVS-Gesture dataset. For each TBR slice, we inject a number of events equal to a slice of noise of length 2.5ms, 5ms, 25ms and 50ms. Spike-TBR still filters the noise out and performs well, even when trained on synthetic random noise.}

\begin{table}
    \centering
    \caption{\rev{We inject in each TBR slice the amount of noise taken from a Prophesee EVK4 equal to 2.5ms, 5ms, 25ms and 50ms. Spike-TBR, although trained on synthetic random noise, still exhibits higher robustness.}}
    \resizebox{.7\columnwidth}{!}{
    \begin{tabular}{l|cccc}
         %\multicolumn{1}{c}{}& \multicolumn{5}{c}{DVSGesture-128} \\ \hline
         Real Noise & 2.5ms & 5ms & 25ms & 50ms \\ \hline
         TBR & 98.10 & 20.83 & 9.09 & 9.09 \\
         Spike-TBR$_{LIF}$ & \textbf{99.24} & \textbf{97.77} & \textbf{84.47} & \textbf{19.69} \\
    \end{tabular}
     }
    \label{tab:realnoise}
\end{table}

\begin{table}[t]
    \centering
    \caption{Effect of the parameter $\beta$ on Spike-TBR$_{LIF}$. Different $\beta$ yield a considerably different sensitivity to noise.}
    \resizebox{\columnwidth}{!}{
    \begin{tabular}{l|ccccc|ccccc}
         \multicolumn{1}{c}{}& \multicolumn{5}{c}{DVSGesture-128} & \multicolumn{5}{c}{DVSLip} \\ \hline
         Noise & 0.0 & 0.005 & 0.01 & 0.03 & 0.05 & 0.0 & 0.005 & 0.01 & 0.03 & 0.05\\ \hline
        $\beta=0.3$ & 98.10 & 96.96 & \textbf{95.36} & \textbf{82.90} & \textbf{19.70} & 45.99 & 11.68  & 3.65 & 0.73 & \textbf{0.73}\\
        $\beta=0.5$ & \textbf{99.60} & \textbf{98.10} & 93.94  & 42.80  & 9.09 & 65.69 & 24.09 & 6.57 & 0.73 & 0.00 \\
        $\beta=0.7$ & \textbf{99.60} & 96.21 & 90.15 & 9.09 & 9.09 & 67.15 & 33.58 & 8.02 & 0.00 & 0.00 \\
        $\beta=0.9$ & \textbf{99.60} & 91.28 & 79.16 & 9.09 & 9.09 & \textbf{75.91} & \textbf{45.99} & \textbf{27.07} & \textbf{2.19} & 0.00\\
    \end{tabular}
     }
    \label{tab:ablations}
\end{table}

\subsection{Effect of Decay Rate}
In all the variants of the LIF neuron, the $\beta$ parameter, defined in Eq.~\ref{eq:beta} plays a decisive role in how noise is handled. In fact, the parameter controls the decay of the membrane potential through time, thus affecting the amount of time the information is retained by the neuron.
In Tab.\ref{tab:ablations} we report an ablation study, varying the parameter in Spike-TBR$_{LIF}$ on two datasets, namely DVSGesture-128 \cite{amir2017low} and DVSLip \cite{tan2022multi}. In particular, we study the effect of $\beta$ on different noise levels.
Interestingly, depending on the dataset, the parameter has an opposite effect: on DVSGesture-128 a small $\beta$ yields a higher resistance to noise, whereas on DVSLip the same is achieved with a high value of $\beta$. The reason for this opposite behavior can be found in the nature of the data. In fact, streams in the DVSLip dataset exhibit a much lower event rate compared to DVSGesture-128, hence relevant information is easily discarded as the membrane decays. On the opposite, DVSGesture-128 videos have a high event rate, which allows the neurons to retain the signal even with small $\beta$, discarding the noise effectively.

% \begin{table}
%     \centering
%     \caption{\rev{Number of spiking and non-spiking operations of Spike-TBR with I3D on average on DVSGesture-128.}}
%     \resizebox{.7\columnwidth}{!}{
%     \begin{tabular}{l|cccc}
%          % \multicolumn{1}{c}{}& \multicolumn{3}{c}{DVSGesture-128} \\ \hline
%           OPs & ACs & MACs & Total & SNN Overhead \\ \hline
%          Spike-TBR & 50.18K & 194.11M & 194.16M & +0.02\%\\
%          %  RecLIF & 50.18K & 193.93M & 193.98M \\
%     \end{tabular}
%      }
%     \label{tab:flops}
% \end{table}

\subsection{Computational Analysis}
\rev{
%To assess the computational footprint of Spike-TBR, we compute the number of operations performed by the network.
We compute SNN-related metrics \cite{chen2023training}: ACs (accumulated computation per synaptic operation only upon the receipt of an incoming spike) and MACs (multiply-accumulate computations per synaptic operation).
The AC operations are those performed by the SNN, the MAC operations are instead performed by the rest of the architecture (I3D). On DVSGesture-128, on average, the model performs 50.18K ACs and 194.11M MACs, thus resulting in a negligible overhead for the SNN (+0.02\%).
}

\section{Conclusions}
We introduced Spike-TBR, a noise-robust event encoding approach that enhances Temporal Binary Representation (TBR) \cite{innocenti2021temporal} by leveraging the inherent temporal dynamics of spiking neural networks.
Our method addresses key limitations of TBR, particularly its sensitivity to noise, by integrating spiking neurons to filter noisy events and preserve critical spatio-temporal information.
Experiments across several datasets demonstrate that Spike-TBR achieves competitive performance on clean data, while it significantly outperforms the traditional approach under noisy conditions. 
The proposed method effectively balances compact event-based representation with enhanced noise robustness, offering a promising avenue for further research in event-driven vision tasks.
\rev{A limitation of the proposed approach, that is worth exploring in future work, includes the fact that the encoding strategy may require fine-tuning when applied to different neuromorphic cameras. Future work will explore training with noise augmentation, evaluating on real-world applications, and extending Spike-TBR to other event-based tasks such as optical flow and object tracking.}

% \begin{table}
%     \centering
%     \caption{.}
%     \resizebox{.9\columnwidth}{!}{
%     \begin{tabular}{l|ccc|ccc}
%          \multicolumn{1}{c}{}& \multicolumn{3}{c}{DVSGesture-128} & \multicolumn{3}{c}{NCALTECH101}\\ \hline
%           OPs & ACs & MACs & Total & ACs & MACs & Total \\ \hline
%          LIF/PLIF & 50.18K & 194.11M & 194.16M & 50.18K & 176.86M & 176.91M \\
%          RecLIF & 50.18K & 193.93M & 193.98M & 50.18K & 176.86M & 176.91M \\
%     \end{tabular}
%      }
%     \label{tab:flops}
% \end{table}

%\bibliographystyle{model2-names}
\bibliographystyle{elsarticle-num-names.bst}
\bibliography{refs}

\end{document}